\documentclass[submission,copyright,creativecommons]{eptcs}
\providecommand{\event}{AREA 2020} 
\usepackage{breakurl}             
\usepackage{underscore}           
\usepackage{cite,xcolor,textcomp,graphicx,algorithmic,amsmath,amssymb,amsfonts,commands,hyperref,dsfont,tabularx,booktabs}
\usepackage[textwidth=3cm,textsize=footnotesize]{todonotes}

\newcommand*{\affaddr}[1]{#1} 
\newcommand*{\affmark}[1][*]{\textsuperscript{#1}}

\title{Statistical Model Checking of \\Human-Robot Interaction Scenarios}

\author{%
	Livia Lestingi\affmark[$\dag$]{\normalfont ,} 
	Mehrnoosh Askarpour\affmark[$*$]{\normalfont ,} 
	Marcello M.Bersani\affmark[$\dag$]{\normalfont ,}  
	Matteo Rossi\affmark[$\dag$]
	\vspace{.15cm}
	\\\affaddr{\affmark[$\dag$] Politecnico di Milano, \texttt{\footnotesize\{firstname\}.\{lastname\}@polimi.it}}
	\vspace{.15cm}
	\\\affaddr{\affmark[*]McMaster University,
	\texttt{\footnotesize askarpom@mcmaster.ca}}
}

\def\titlerunning{Statistical Model Checking of Human-Robot Interaction Scenarios}
\def\authorrunning{L. Lestingi, M. Askarpour \& M.M. Bersani \& M. Rossi}

\begin{document}
\maketitle

\begin{abstract}
Robots are soon going to be deployed in non-industrial environments. Before society can take such a step, it is necessary to endow complex robotic systems with mechanisms that make them reliable enough to operate in situations where the human factor is predominant. 
This calls for the development of robotic frameworks that can soundly guarantee that a collection of properties are verified at all times during operation. 
While developing a mission plan, robots should take into account factors such as human physiology. 
In this paper, we present an example of how a robotic application that involves human interaction can be modeled through hybrid automata, and analyzed by using statistical model-checking. 
We exploit statistical techniques to determine the probability with which some properties are verified, thus easing the state-space explosion problem. 
The analysis is performed using the Uppaal tool.
In addition, we used Uppaal to run simulations that allowed us to show non-trivial time dynamics that describe the behavior of the real system, including human-related variables.
Overall, this process allows developers to gain useful insights into their application and to make decisions about how to improve it to balance efficiency and user satisfaction.
\end{abstract}

\section{Introduction}
\label{sec:intro}
Robots have been mostly used in industrial environments. Indeed, manufacturing lines that feature robotic manipulators, as well as mobile robots in automated warehouses, are very common nowadays. Nevertheless, these applications usually involve strong workspace separation between humans and robots (e.g., through cages or dedicated lanes). Over the last few years, thanks to the phenomenon called Industry 4.0, a lot of effort has been put into trying to inject human-robot collaboration into industrial contexts. 
On the other hand, robots employed in non-industrial settings still represent extraordinary cases despite their massive potential. Their deployment could, for example, relieve health professionals from the most repetitive tasks, such as transporting objects or delivering paperwork, so that they would have more time for duties which are exclusive to humans, i.e., taking care of patients.  Similarly, robots could provide support in disaster relief situations, for instance when buildings at risk of collapse need to be inspected.
\\
Before users can welcome personal assistant robots into their daily routines, they require strong guarantees about the fact that these machines will not cause any harm nor any damage. The issue of safety in service robotics has already been tackled in previous works, firstly by standards, as assessed by Virk et al. \cite{virk2008iso} (most notably ISO:13482 \cite{ISO:13482}). The challenges of minimizing risk for service robots, as argued by Jacobs et al. \cite{jacobs2014iso}, are due to the diversity of users that this technology reaches, and to the fact that most applications imply intended contact. \\
However, safety is not the only priority. It is also paramount that users perceive the addition of robots as an improvement rather than a step back. This involves building trust between the human and the machine (e.g., by making the robot able to transparently share its status), and assuring that the robot plan does not prevail over human needs. This is especially important when users in delicate conditions are involved, such as patients who may be too tired or in discomfort to keep up with the robot's actions.\\
We have selected a relevant scenario where human-robot interaction is the most relevant aspect to be considered. A human and a battery-powered mobile robot are the main elements of the scenario. The novelty lies in the fact that non-traditional variables pertaining to human mental and physical conditions are embedded in the model. Given the complex dynamics of the system, the most appropriate choice is to develop the formal model as a network of Hybrid Automata.
The modeling tool is Uppaal \cite{behrmann2004tutorial} and its extension Uppaal SMC \cite{david2015uppaal}. Statistical Model Checking (SMC) assesses the probability of occurrence of certain critical events based on time-bounded runs of the system. In particular, in this paper we explore the likelihood of the human reaching full exhaustion while the robot is still moving: this highlights the need of robot decision-making policies that are aware of human needs.
In the future, we envision the development of features that make the scenario easily customizable and verifiable by users who possess a technical background though not on formal methods. This could prove substantial, for example, to healthcare specialists willing to estimate the outcome of a real application case.\\
The rest of the paper is organized as follows: Section \ref{sec:rw} surveys related works existing in literature; Section \ref{sec:bg} outlines the background of the work; Section \ref{sec:scenario} introduces in detail the selected case study and Section \ref{sec:ha} presents the main contribution of the paper, i.e. the hybrid automata; Section \ref{sec:exp} reports the experimental results; finally, Section \ref{sec:concl} draws some conclusion.
%
\section{Related Work}
\label{sec:rw}
%
The literature features a number of works on the use of timed automata to model complex systems. Molnar et al. \cite{molnar2011hybrid} present a layered modeling architecture where agents correspond to hybrid systems interacting with each other.
Wang et al. \cite{wang2018formal}\cite{li2017formal} explore the possibility of modeling a simple robotic application using the Uppaal tool. They model every robot component as a set of automata that synchronize with each other to perform the mission. Once the model formally satisfies a set of properties, they automatically generate executable C++ code.
Similarly, Halder et al. \cite{halder2017formal} model a mobile Kobuki robot ROS-based application in Uppaal, and subsequently perform model-checking to verify relevant properties about communication between nodes, e.g., that loss of information never occurs.
The approach presented by Aniculaesei et al. \cite{aniculaesei2016towards} shows an approximation of the environment by means of static and dynamic obstacles, that might obstruct the movement of a mobile robot. In this case, Uppaal is used to verify collision-avoidance conditions.
The work by Zhou et al. \cite{zhou2016timed} is centered on motion planning, where the mission is specified with an MITL formula from which an automaton is generated to find feasible runs. 
%
%
Model-checking techniques are also applied to formal models of robotic systems.
Webster et al. \cite{webster2014formal} analyze the case study of a personal robotic assistant operating in a house-simulating laboratory. High-level rules are translated into a Brahms workframe, which is subsequently translated into the language used by the SPIN model-checker. 
The work by Arai and Schlingloff \cite{arai2017model} implements statistical model checking to evaluate the performances of autonomous transport robots in an industrial plant. In their case, the verification process aims at finding the optimal value of certain system parameters, e.g., how many robots are needed to complete the mission. Despite the intrinsic limitations of simulation, they manage to obtain results exhaustive enough so that stakeholders can make their decisions.
Statistical model checking is also pivotal in the works by Foughali et al. \cite{foughali2017toward}\cite{foughali2019statistical}. 
The authors compare different formalisms to model and formally verify concurrency \cite{foughali2017toward}. In particular, they select Timed Transition Systems to formalize such components and perform statistical verification of the model of a quadricopter \cite{foughali2019statistical}. In this case, relevant properties involve the machine's capability to respond to requests in an orderly and consistent fashion to avoid accidents.\\
%
Despite model-checking receiving some attention over the last few years, most of the works applying formal verification approaches to robotics tend to focus on the robot itself and its internal structure. In particular, formal verification mostly deals with internal safety and integrity properties. Verifying this type of property holds major importance, for example, when the product is undergoing the certification process to be released into the market. On the other hand, applications that heavily feature human-interaction also require higher-level models and the verification of properties of a different nature. In this paper, we present an example of this type of model that mostly focuses on human-related aspects.
%

\section{Background}
\label{sec:bg}
In this Section, we introduce the bases on which the model of the scenario has been built. 
The selected formalism, as mentioned in Section \ref{sec:intro}, is that of \emph{hybrid automata}. 
Timed automata are built on the simplest time dynamics \cite{furia2012modeling}: each model features specific variables called \emph{clocks}, whose value increases linearly with time unless they are reset. Hybrid automata \cite{alur1995algorithmic} can be considered an extension of the aforementioned formalism since they support generic time dynamics. 
As per the semantics defined by \cite{alur1995algorithmic}, each automaton is composed of a set of \emph{locations}.  Each location can be endowed with constraints, called \emph{flow conditions}, that determine how variables evolve with time. 
The switch from one location to another is realized through \emph{transitions}. If a transition is endowed with a \emph{guard} condition, it may be fired only when the condition is true. In some cases, it is useful to model transitions that only fire once an event $e$ occurs: these are labeled by $e!$ and $e?$ if the transition respectively triggers the event $e$ or awaits its occurrence. These labels are also called \emph{channels} \cite{behrmann2004tutorial} since they allow the synchronization of concurrent systems in a way similar to the publish/subscribe pattern.\\
Hybrid automata are useful to model systems whose dynamics can be described by a set of non-elementary ODEs (Ordinary Differential Equations). In our specific case, ordinary timed automata would not have proven sufficiently expressive to capture either robot or human behaviors that possess complex dynamics. For example, hybrid automata are well-suited to capturing human-related variables in the system, such as \his{} fatigue. The adopted model of human fatigue $F(t)$ is the one proposed by Konz \cite{konz2000work}\cite{jaber2013incorporating}, which can also be found in eq.\eqref{eq:fatigue}.
\begin{equation}
\vspace{-0.1cm}
\label{eq:fatigue}
F(t) =
	\begin{cases}
		1 - e^{-\lambda t} & \mbox{(a)} \\
		e^{-\mu t} & \mbox{(b)}
	\end{cases}
\end{equation}
Eq.\eqref{eq:fatigue}$(a)$ represents how the level of fatigue increases (to a maximum value of $F=1$) with time $t$ while walking, and eq.\eqref{eq:fatigue}$(b)$ how it decreases with time while resting ($F=0$ means full recovery). Parameters $\lambda$ and $\mu$ determine the duration of full fatigue accumulation/recovery cycles: if they have a low value, fatigue accumulation / recovery will be slow \cite{jaber2013incorporating}. Finally, if $F(t_{max})=1$, $t_{max}$ is called Maximum Endurance Time (MET) \cite{jaber2013incorporating}.
%
As for the robot, we assume it is powered by a lithium battery with a typical charge/discharge profile \cite{tremblay2007generic}. The battery charge is set to $100\%$ at the beginning of the simulation, then it decays rapidly until $80\%$ (with rate $r_1$) and between $20\%$ and $0\%$ (rate $r_3$), whereas the discharge rate decreases significantly in the nominal zone (rate $r_2$), i.e., between $80\%$ and $20\%$.
%

\section{Human-Robot Interaction Scenario}
\label{sec:scenario}
%
\begin{figure}[t]
\centering 
\includegraphics[width=.9\columnwidth]{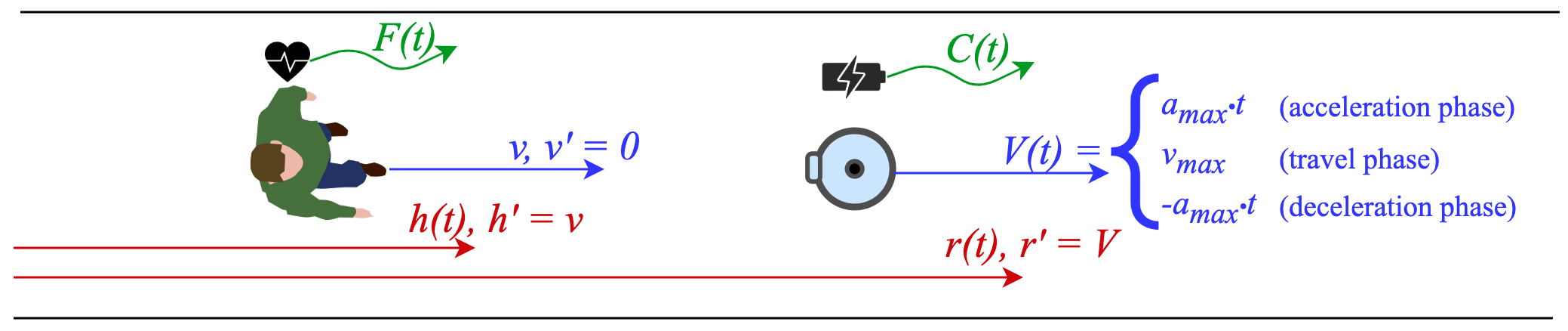}\\
\caption{\footnotesize Diagram representing the scenario: green arrows $\mathit{F(t)}$ and $\mathit{C(t)}$ represent human fatigue and battery charge; red arrows $\mathit{h(t)}$ and $\mathit{r(t)}$ are the distances covered respectively by human and robot; blue arrows represent the constant speed of the human $\mathit{v}$ and the robot's speed $\mathit{V(t)}$ which follows a trapezoidal profile.} 
\label{fig:scheme}
\vspace{-0.25cm}
\end{figure}
The case study involves a robot in a healthcare setting leading a newly registered patient to \his{} room or a doctor's office. The actors, also depicted in Figure \ref{fig:scheme}, are a battery-powered mobile robot and a human, and the synchronization is realized by having the robot lead the human along a hallway.\\ 
In this scenario, the robot can start moving, stop, and start recharging. We also assume that the robot can recharge regardless of its position, and stop recharging when the battery charge is back to its full. In real life, the decision to start and stop could be caused by several factors, such as a sudden obstacle, which are not captured by the model. The simplest way to simulate this behavior is to assume that the probability to leave the \emph{idle} state has an exponential distribution $1-e^{-\lambda t}$. As a consequence, the robot is more likely to take action as time passes and quicker as $\lambda$ increases.\\
Since we are focusing on the interaction between the human and the robot rather than between the agents and the environment, assuming that the hallway has infinite length does not undermine the efficacy of the model. 
%
For the human, we assume \he{} walk with constant speed $v$. The robot, instead, displays a trapezoidal velocity profile, typical in robotic applications \cite{chettibi2006suboptimal}, which identifies three phases: the acceleration phase, where acceleration is constant and velocity ($V$ in Figure \ref{fig:scheme}) increases linearly with time; the travel phase, where acceleration is $0$ and velocity is constantly equal to its maximum; the deceleration phase, equivalent to the first phase but with negative acceleration. 
Human fatigue and battery charge ($F$ and $C$ in Figure \ref{fig:scheme}) are modeled as explained in Section \ref{sec:bg}.\\
In this setting, with the specified boundaries, some critical events can occur that require verification. The aspect that we are most interested in analyzing is the likelihood of the human passing out due to over-exhaustion while the robot is moving, thus causing the failure of the mission.
%
\section{Hybrid Automata Model}
\label{sec:ha}
\begin{figure}[t]
\centering 
\includegraphics[width=.9\columnwidth, trim=0 0.5cm 0 0]{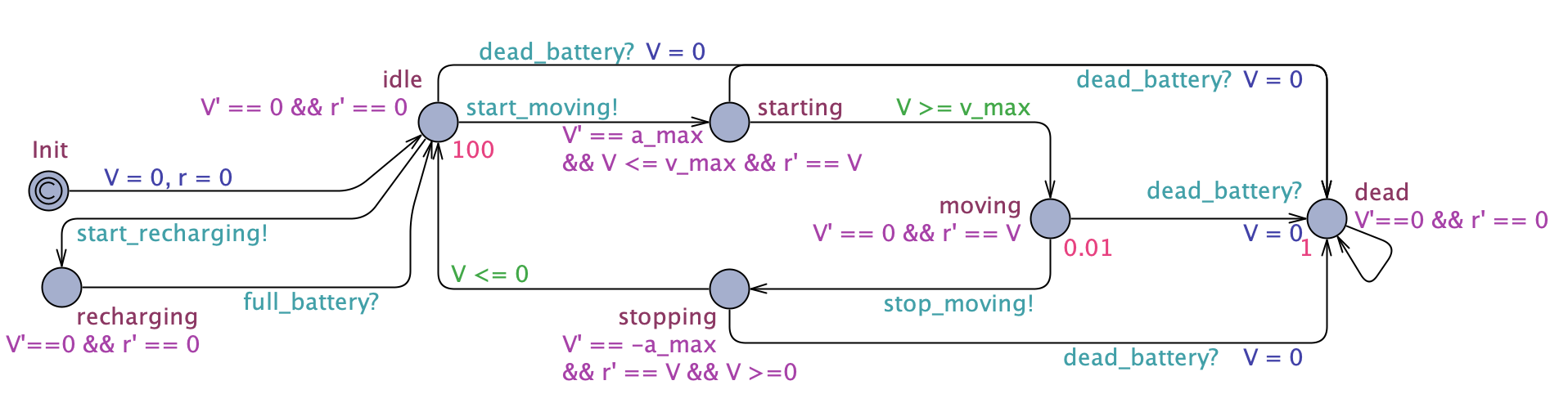}\\
(a)\\
\includegraphics[width=.9\columnwidth, trim=0 0.5cm 0 0]{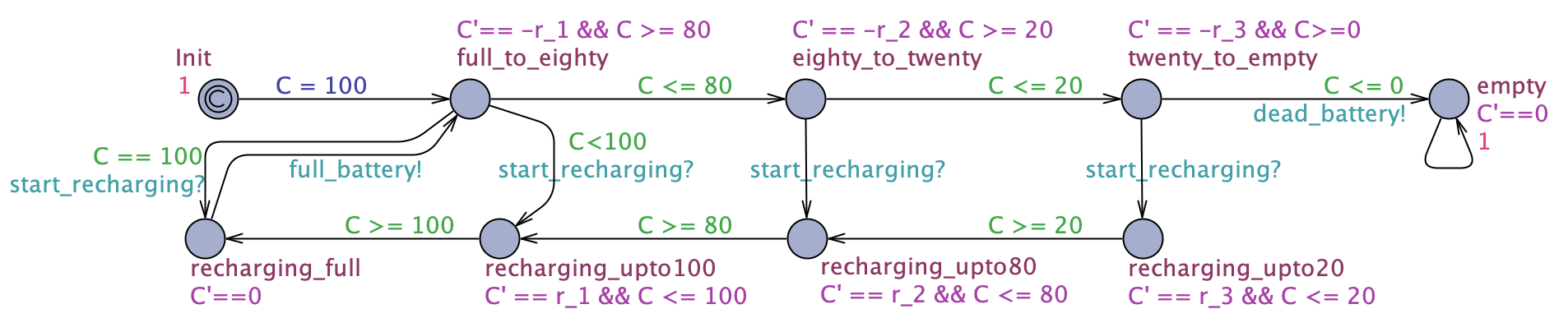}\\
(b)\\
\includegraphics[scale=0.4, trim=0 0.5cm 0 0cm]{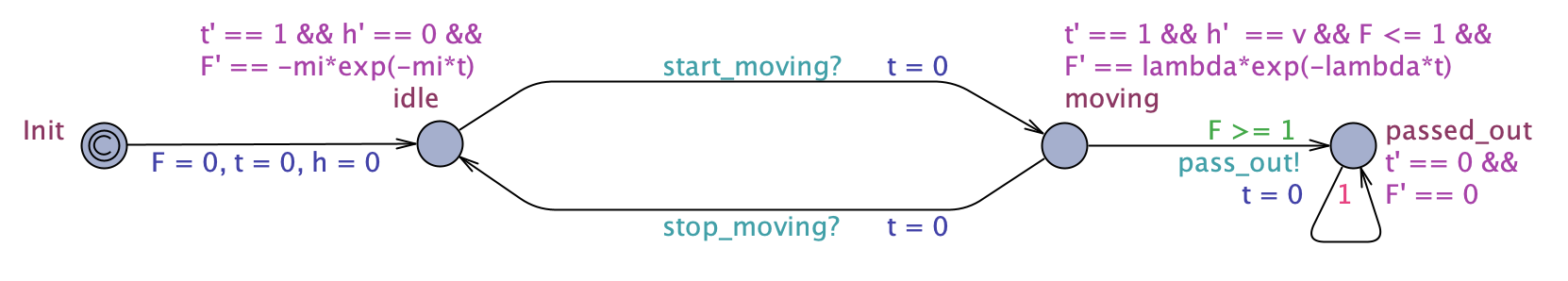}\\
(c)\\
\caption{\footnotesize Hybrid Automata modeling: (a) robot, (b) robot battery, (c) the human.}
\label{fig:ha}
\vspace{-0.5cm}
\end{figure}
The complete model of the agents described in Section \ref{sec:scenario} is composed of the three hybrid automata depicted in Figure \ref{fig:ha}, which will now be presented in detail. \\
The automaton modeling the robot's behavior can be found in Figure \ref{fig:ha}(a). Variables $r$ and $V$ model the distance covered by the robot and its velocity described in Section \ref{sec:scenario}, while constants $\mathsf{a_{max}}$ and $\mathsf{v_{max}}$ (also in Figure \ref{fig:scheme}) correspond to maximum acceleration and velocity. 
In the initial state $\mathit{idle}$, the robot is not moving, hence its flow condition is $\dot{V}=0, \dot{r}=0$. From the initial state, the robot can fire two transitions, triggering two events: $\mathtt{start\_recharging}$ and $\mathtt{start\_moving}$. In this preliminary version of the model, the choice is non-deterministic. 
In the first case, the automaton enters location $\mathit{recharging}$ and the flow conditions remain unvaried. This location can be left only when the event $\mathtt{full\_battery}$ is triggered. 
In the second case, the robot starts moving, hence the velocity profile described in Section \ref{sec:scenario} is initiated. The automaton first enters the location $\mathit{starting}$, \ie{} the acceleration phase, where the following holds: $\dot{V}=\mathsf{a_{max}}, \dot{r}=V$. When guard $V\geq \mathsf{v_{max}}$ is true, the robot switches to location $\mathit{moving}$, corresponding to the travel phase, where $\dot{V}=0, \dot{r}=\mathsf{v_{max}}$. 
When the robot fires the event $\mathtt{stop\_moving}$, it switches to location $\mathit{stopping}$, corresponding to the deceleration phase, hence $\dot{V}=-\mathsf{a_{max}}, \dot{r}=V$. This phase is completed when guard $V\leq 0$ is satisfied, thus the robot returns to its $\mathit{idle}$ state. 
Locations $\mathit{starting}$ and $\mathit{stopping}$ are endowed with invariants $V\leq \mathsf{v_{max}}$ and $V\geq 0$, so that the outgoing transitions are respectively fired when $V$ is exactly equal to $\mathsf{v_{max}}$ and $0$. 
All locations, except for $\mathit{recharging}$, have an additional outgoing transition towards the deadlock state $\mathit{dead}$, which fires when event $\mathtt{dead\_battery}$ occurs. In this state, the robot is forced to stop, hence $V$ is reset to $0$ and $\dot{V}=0, \dot{r}=0$.\\
The automaton corresponding to the battery, in Figure \ref{fig:ha}(b), captures the charge/discharge cycles described in Section \ref{sec:bg}. 
The time-dependent variable $C$ models the level of charge, which is set to $100$ in the initial state $\mathit{full\_to\_80}$. The three discharge cycle phases are identified by as many locations ($\mathit{full\_to_\_80}$, $\mathit{80\_to\_20}$, and $\mathit{20\_to\_empty}$). There is a deadlock state $\mathit{empty}$ that captures the case in which the battery is fully discharged and human intervention is required to make the robot operational again. 
The switch from one location to the next takes place when $C=\mathsf{C_{th}}$, with $\mathsf{C_{th}}$ respectively equal to $80, 20$, and $0$. This is achieved through the invariant $C\geq \mathsf{C_{th}}$ on the starting location and the guard $C\leq \mathsf{C_{th}}$ on the outgoing transition. 
The last transition from $\mathit{20\_to\_empty}$ triggers the event $\mathtt{dead\_battery}$, causing also the robot automaton to enter its deadlock state. 
The time-dynamics of $C$ are constrained by flow conditions on the above-mentioned locations, respectively $\dot{C}=-\mathsf{r_1}, \dot{C}=-\mathsf{r_2}, \dot{C}=-\mathsf{r_3}$ and $\dot{C}=0$ in the $\mathit{empty}$ state. 
The locations corresponding to the recharge cycle ($\mathit{recharging\_upto20}$, $\mathit{recharging\_upto80}$, $\mathit{recharging\_upto100}$ and $\mathit{recharging\_full}$) have a dual behavior with flow conditions $\dot{C}=\mathsf{r_3}, \dot{C}=\mathsf{r_2}, \dot{C}=\mathsf{r_1}$ and $\dot{C}=0$. While recharging, the passage from one location to the next takes place when $C=\mathsf{C_{th}}$ with: $\mathsf{C_{th}}$ equal to $20, 80, 100$, invariants in the form $C\leq \mathsf{C_{th}}$ and guard conditions $C\geq \mathsf{C_{th}}$. 
When location $\mathit{recharging\_full}$ is reached, the \emph{urgent} channel \cite{behrmann2004tutorial} $\mathtt{full\_battery}$ is immediately triggered, so that the robot automaton also leaves its $recharging$ location. The transition from the location of a discharging phase to its corresponding recharging one fires when the robot triggers the event $\mathtt{start\_recharging}$.\\
Finally, Figure \ref{fig:ha}(c) shows the automaton that models the human. 
In this case, the variables with non-trivial time-dynamics are the covered distance $h$ and the fatigue $F$, whose model has already been described in Section \ref{sec:bg}. A clock $t$, whose value grows linearly with time unless it is reset, is also required to keep track of active and resting time frames duration. 
When the automaton is in its initial state $\mathit{idle}$, we assume the human is in its resting phase, hence the flow conditions are $\dot{F}=-\mu e^{-\mu t}, \dot{h}=0$. 
When the robot triggers the event $\mathtt{start\_moving}$, the human switches to its $\mathit{moving}$ location and the clock $t$ is also reset. In this case, since fatigue increases with time, the flow condition imposes $\dot{F}=\lambda e^{-\lambda t}$ and $\dot{h}=v$, as the distance increases as the human follows the robot. This location is also endowed with the invariant $F\leq 1$: once this condition becomes false, it means the fatigue has reached the maximum tolerable value and the human enters its deadlock state $\mathit{passed\_out}$, where $\dot{F}=0$. The switch from location $\mathit{moving}$ back to $\mathit{idle}$ occurs when the robot triggers the event $\mathtt{stop\_moving}$ and the human goes back to rest.
%
\section{Experimental Results}
\label{sec:exp}
%
\subsection{Verification and Simulation}
The hybrid automata presented in Section \ref{sec:ha} have been implemented in the Uppaal tool \cite{behrmann2004tutorial} and subjected to SMC experiments \cite{david2015uppaal}. Given the parametric model, we assign the values justified in the following. Having established that $1$ time unit corresponds to $1s$ and choosing the TurtleBot2\footnote{https://www.turtlebot.com/turtlebot2/} specifications as reference, we set $\mathsf{v_{max}} = 65cm/s$ and $\mathsf{a_{max}} = 50cm/s^2$. We also assume that the human moves at the same speed, although lower than \his{} average, since \he{} need to follow the robot ($\mathsf{v} = 65cm/s$).
To comply with a real mobile robot charge life, the chosen battery parameters are $\mathsf{r_1} = 0.035, \mathsf{r_2} = 0.008, \mathsf{r_3} = 0.055$, so that the resulting battery life is approximately $2.5h$. Finally we set $\lambda = \mu = 0.005$ (eq.\ref{eq:fatigue}), which would mean full recovery/exhaustion after about $15min$ of resting/walking. \\
\begin{figure}[t]
\centering 
\includegraphics[width=\columnwidth, trim=0 0.6cm 0 0]{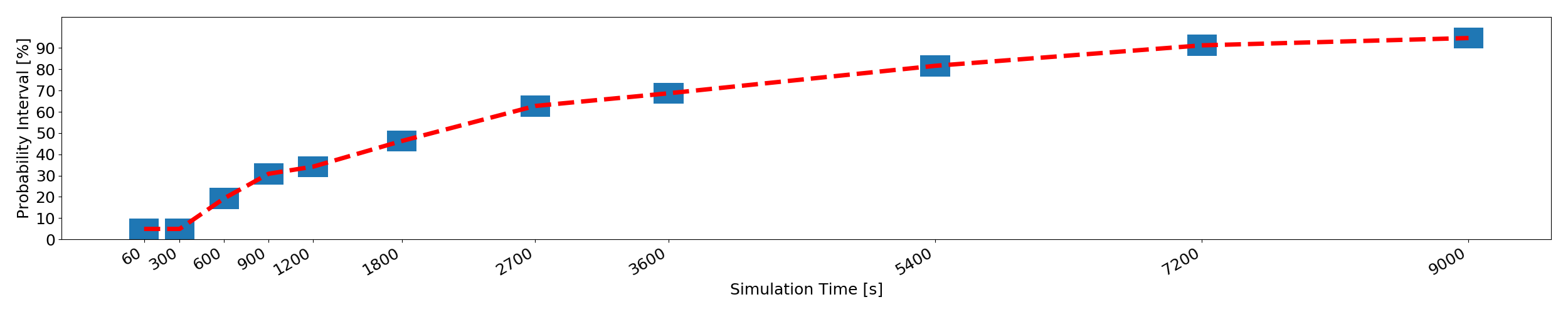}\\
(a)\\
\includegraphics[width=\columnwidth, trim=0 0.6cm 0 0]{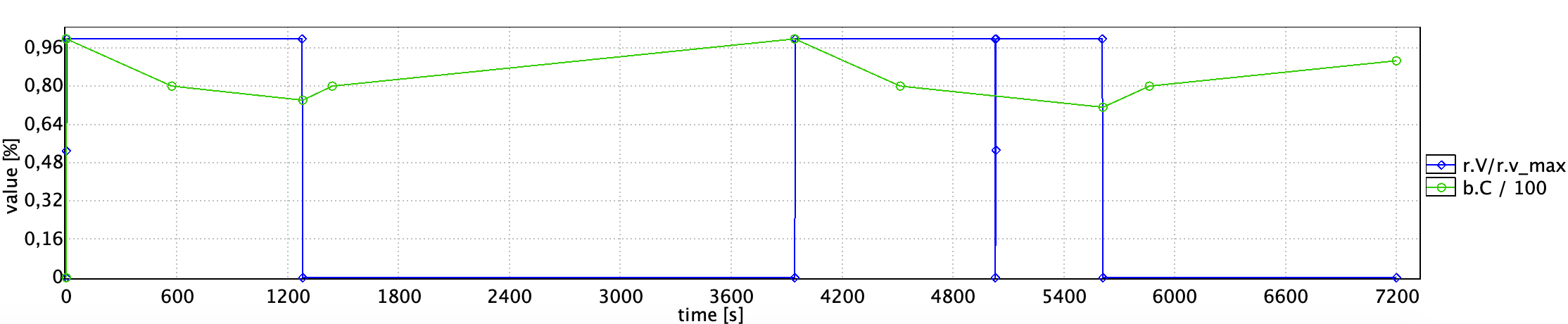}\\
(b)\\
\includegraphics[width=\columnwidth, trim=0 0.6cm 0 0]{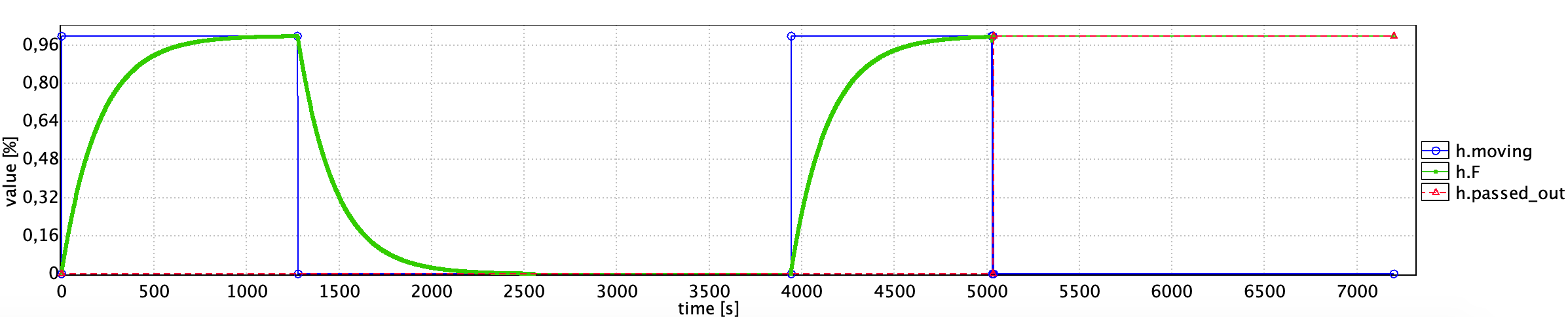}\\
(c)\\
\vspace{-0.2cm}
\caption{\footnotesize (a) shows the probability ranges (blue bars) calculated by Uppaal over increasing values of time, while the red line connects the average points of such intervals; (b) and (c) provide an example of system simulation in which the mission fails (the human passes out after about $5000s$): in (b) we show robot-related variables, and in (c) human-related ones.} 
\label{fig:exp1}
\vspace{-0.2cm}
\end{figure}
As mentioned in Section \ref{sec:scenario}, the property to verify, formally expressed in eq.\ref{eq:passoutpr}, is that the automaton representing the human will \textit{eventually} ($\diamond$) enter location $passed\_out$ while the robot is moving, hence while the automaton is either in location $starting$ or $moving$. Classical model-checking properties such as \emph{deadlock} are subsumed by \ref{eq:passoutpr}.
\vspace{-0.2cm}
\begin{equation}
\label{eq:passoutpr}
\large\mathit{P}_{\leq t_s}[\diamond\enspace passed\_out \land (starting \lor moving)]
\vspace{-0.2cm}
\end{equation}
The tool allows us to evaluate the probability of the property being verified in the bounded interval $\left[0, t_s\right]$, which determines the timespan of simulations. By running the verification \footnote{On a MacOs 10.14 machine with 8GB of RAM and 2 cores, running the whole experiment takes about 10min.} for monotonically increasing values of $t_s$, we analyse how the probability evolves with time. \\
Figure \ref{fig:exp1}(a) shows that the probability of the patient passing out is negligible for the first $5min$, it increases with time and finally crosses the $90\%$ threshold after $2h$. This outcome is only apparently inconsistent with the $15min$ Maximum Endurance Time (MET). Indeed, this value of MET refers to intervals of non-stop walking, whereas in our model the robot can stop during operation and this gives the patient time to rest. An example of simulation is given in Figure \ref{fig:exp1}(b) and (c), where the patient is fully exhausted at around time $5000$ after walking for about $16min$ without stopping. The simulation also gives us some insights on the robot behavior and its battery, whose charge (the green line in Figure \ref{fig:exp1}(b)) coherently decreases during movement and increases when the robot stops moving and starts recharging.\\
%
\subsection{Beyond Verification}
Were stakeholders to synthesize a robot controller for this case study, the results of this experiment should be taken into account by adding a fatigue monitoring loop as a decision-making layer for the robot. 
The core idea is to trigger the events $\mathtt{start\_moving}$ and $\mathtt{stop\_moving}$ only when the value of fatigue belongs to an acceptable range: if the pair is moving and the patient becomes excessively tired, the robot should stop and restart when the human has had time to rest.
In reality, this could be achieved by monitoring the distance between the robot and human: if this value crosses a certain threshold than the human is likely not being able to keep up. Alternative solutions could involve the use of a wearable device that can monitor the heartbeat or machine learning techniques to estimate human fatigue in real-time \cite{gordienko2017deep}.
%
\section{Conclusions}
\label{sec:concl}
We have presented a high-level modeling approach of a human-robot interaction case study. By building upon these bases, we intend to add to the model a robot controller that takes into account non-traditional human-related factors. Further future refinements include additional interaction patterns and a realistic floor plan layout. Once the framework has reached a suitable stage of development, it will be tested in a real environment to verify the effectiveness of the synthesized controllers. 

\bibliographystyle{eptcs}
\bibliography{references}
\end{document}